\documentclass[sigconf]{acmart}

\AtBeginDocument{%
  }

\setcopyright{acmlicensed}
\copyrightyear{2025}
\acmYear{2025}
\acmDOI{XXXXXXX.XXXXXXX}

\acmConference[Sensys'25]{The 23rd ACM Conference on Embedded Networked Sensor Systems}{May 06--09, 2025}{Irvine, USA}
\acmBooktitle{The 23rd ACM Conference on Embedded Networked Sensor Systems (SenSys'25), May 06--09, 2025, Irvine, USA}
\acmPrice{15.00}
\acmISBN{978-1-4503-XXXX-X/18/06}

\usepackage{booktabs} 
\usepackage{balance} 
\usepackage{graphicx}
\usepackage{url}
\usepackage{caption}
\usepackage{epstopdf}
\usepackage{bm}
\usepackage{subcaption}
\usepackage{pdfpages}
\usepackage{capt-of}
\usepackage{lipsum}
\usepackage{wrapfig}
\usepackage{pifont}
\usepackage{tabularx}
\usepackage{hyperref}
\usepackage{epsfig}
\usepackage{xcolor}
\usepackage{makecell}
\usepackage{multirow}
\usepackage{nomencl}
\usepackage{xspace}
\usepackage{amsmath}
\usepackage{mathtools}
\usepackage{siunitx} 
\usepackage{enumitem}
\usepackage{afterpage}

\definecolor{mygreen}{RGB}{55, 192, 120}
\definecolor{myred}{RGB}{230, 90, 96}

\usepackage[capitalize]{cleveref}
\Crefname{section}{Sec.}{Secs.}
\Crefname{section}{Section}{Sections}
\Crefname{table}{Tab.}{Tabs.}
\Crefname{table}{Table}{Tables}

\PassOptionsToPackage{pagebackref,breaklinks,colorlinks}{hyperref}
\usepackage{hyperref}
\def\onedot{.\xspace}
\def\eg{\emph{e.g}\onedot} 

\def\ie{\emph{i.e}\onedot} 

\def\cf{\emph{cf}\onedot}

\def\wrt{w.r.t\onedot}

\DeclareMathOperator*{\argmin}{arg\,min}

\newcommand{\sysname}{\texttt{ThermoHands}\xspace}

\definecolor{mygreen}{RGB}{55, 192, 120}
\definecolor{myred}{RGB}{230, 90, 96}
\title{ThermoHands: A Benchmark for 3D Hand Pose 
Estimation from Egocentric Thermal Images}

\author{Fangqiang Ding}
\authornote{Equal contribution.}
\affiliation{%
  \institution{University of Edinburgh}
  \city{Edinburgh}
  \country{United Kingdom}}
\email{f.ding-1@sms.ed.ac.uk}

\author{Yunzhou Zhu}
\authornotemark[1]
\affiliation{%
  \institution{Georgia Institute of Technology}
  \city{Atlanta}
  \country{USA}}
\email{lawrencezhu@gatech.edu}

\author{Xiangyu Wen}
\affiliation{%
  \institution{University of Edinburgh}
  \city{Edinburgh}
  \country{United Kingdom}}
\email{wenxiangyu2001@gmail.com}

\author{Gaowen Liu}
\affiliation{%
  \institution{Cisco Research}
  \city{San Francisco}
  \country{USA}}
\email{gwliu213@gmail.com}

\author{Chris Xiaoxuan Lu}
\authornote{Corresponding author: xiaoxuan.lu@ucl.ac.uk}
\affiliation{%
  \institution{University College London}
  \city{London}
  \country{United Kingdom}}
\email{xiaoxuan.lu@ucl.ac.uk}

\renewcommand{\shortauthors}{F. Ding et al.}

\begin{abstract} 
    Designing egocentric 3D hand pose estimation systems that can perform reliably in complex, real-world scenarios is crucial for downstream applications. Previous approaches using RGB or NIR imagery struggle in challenging conditions: RGB methods are susceptible to lighting variations and obstructions like handwear, while NIR techniques can be disrupted by sunlight or interference from other NIR-equipped devices. To address these limitations, we present ThermoHands, the first benchmark focused on thermal image-based egocentric 3D hand pose estimation, demonstrating the potential of thermal imaging to achieve robust performance under these conditions. The benchmark includes a multi-view and multi-spectral dataset collected from 28 subjects performing hand-object and hand-virtual interactions under diverse scenarios, accurately annotated with 3D hand poses through an automated process. We introduce a new baseline method, TherFormer, utilizing dual transformer modules for effective egocentric 3D hand pose estimation in thermal imagery. Our experimental results highlight TherFormer's leading performance and affirm thermal imaging's effectiveness in enabling robust 3D hand pose estimation in adverse conditions.
\end{abstract}


\begin{document}
\maketitle

\begin{figure*}[!tbp]
    \includegraphics[width=\textwidth]{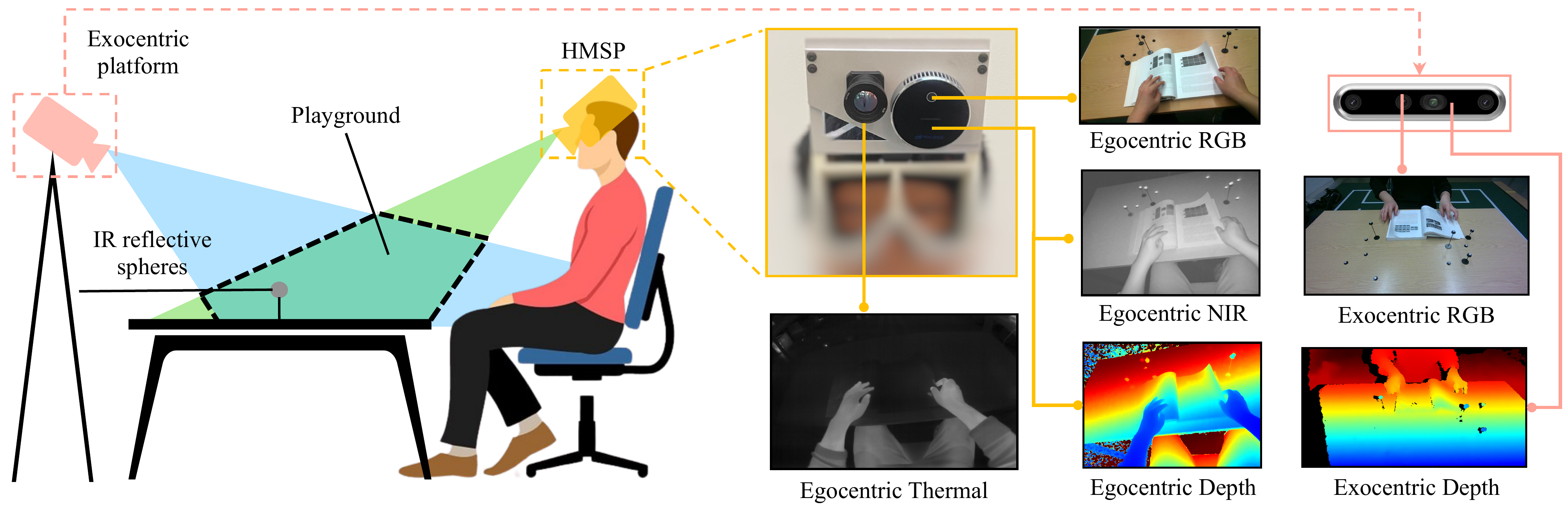}
    \caption{{Data capture setup with the customized head-mounted sensor platform (HMSP) and exocentric platform recording multi-view multi-spectral images of two-hand actions performed by participants. }}
    \label{fig:capture}
\end{figure*}

\section{Introduction}
Egocentric 3D hand pose estimation is critically important for interpreting hand gestures across various applications, ranging from extended reality (XR)~\cite{wu2020hand,sagayam2017hand,marchand2015pose,liang2015ar}, to human-robot interaction~\cite{sampieri2022pose,gao2021dynamic,gao2019dual}, and to imitation learning~\cite{qin2022dexmv,chen2023bi,antotsiou2018task}. Its importance has been magnified with the advent of advanced XR headsets such as the Meta Quest series~\cite{meta_quest} and Apple Vision Pro~\cite{apple_vision_pro_2024}, where it serves as a cornerstone for spatial interaction and immersive digital experiences.

While current research of hand pose estimation primarily focuses on RGB image-based methods~\cite{lin2021two,li2022interacting, wen2023hierarchical, Fu_2023_ICCV,cho2023transformer}, these approaches are particularly vulnerable to issues related to lighting variation and occlusions caused by handwear, \eg, gloves or large jewellery~\cite{youtube_video_2024, apple_support_gloves_2024}. These challenges underscore the imperative for robust egocentric 3D hand pose estimation capable of performing reliably in a variety of common yet complex daily scenarios. 
The prevailing approach to facilitate robust hand pose estimation in low-light conditions utilizes \emph{near infrared} (NIR) cameras paired with active NIR emitters. This technology, invisible to the human eye, leverages active NIR emitter-receiver configurations for depth estimation through time-of-flight (ToF) or structured lighting. Nevertheless, active NIR systems are more power-intensive compared to passive sensing technologies~\cite{IntelRSL515, IntelRealSense2023D455} and are prone to interference from external NIR sources, such as sunlight~\cite{suarez2012using} and other NIR-equipped devices~\cite{seewald2019toward}. Consequently, these vulnerabilities restrict the effectiveness of hand pose estimation under bright daylight conditions and in situations where multiple augmented reality (AR) or virtual reality (VR) systems are used for collaborative works. 




In contrast to NIR-based methods, thermal imaging cameras offer a passive sensing solution for hand pose estimation by capturing long-wave infrared (LWIR) radiation emitted from objects, thereby eliminating reliance on the visible light spectrum~\cite{lloyd2013thermal}. This unique attribute of thermal imaging introduces several benefits for hand pose estimation. Primarily, it accentuates the hand's structure via temperature differentials, negating the effects of lighting variability. Moreover, thermal cameras can detect hands even under handwear such as gloves by identifying heat transmission patterns. This ability ensures a stable and consistent representation of hands, independent of any coverings, thereby broadening the scope and reliability of hand pose estimation across various scenarios. 


Building on the above insights, this study probes the following research question: \emph{Can egocentric thermal imagery be effectively used for 3D hand pose estimation under various conditions (such as different lighting and handwear), and how does it compare to techniques using RGB, NIR, and depth\footnote{For readability, we treat depth and NIR as two `spectra', despite their usual overlap.} spectral imagery?} 
To answer this, we introduce \sysname\footnote{Project page: \url{https://thermohands.github.io/}.}, the first benchmark specifically tailored for egocentric 3D hand pose estimation utilizing thermal imaging. This benchmark is supported by a novel multi-spectral and multi-view dataset  
designed for egocentric 3D hand pose estimation and is unique in comprising thermal, NIR, depth, and RGB images (\cf~\cref{fig:capture}). Our dataset emulates real-world application contexts by incorporating both hand-object and hand-virtual interaction activities, with participation from 28 subjects to ensure a broad representation of actions. To offer a thorough comparison across spectral types, we gather data under five distinct scenarios, each characterized by varying environments, handwear, and lighting conditions (\cf~\cref{tab:stat}). Considering the challenges associated with manually annotating large-scale 3D hand poses, we developed an automated annotation pipeline. This pipeline leverages multi-view RGB and depth imagery to accurately and efficiently generate 3D hand pose ground truths through optimization based on the MANO model~\cite{MANO} (\cf~\cref{fig:anno}).

Together with the multi-spectral dataset, we introduce a new baseline method named \emph{TherFormer}, specifically designed for thermal image-based egocentric 3D hand pose estimation (\cf~\cref{fig:network}). This approach is notable for its two consecutive transformer modules, \ie, mask-guided spatial transformer and temporal transformer, which encode spatio-temporal relationship for 3D hand joints without losing the computation efficiency.

Our validation process begins with verifying the annotation quality, which averages an accuracy of 1cm (\cf~\cref{tab:annotation}). We then benchmark \emph{TherFormer} against leading methods (\cf~\cref{tab:results}) and compare the performance of various spectral images (\cf~\cref{tab:spectra}, \cref{tab:challenge} and \cref{fig:challenge}). The findings underscore thermal imagery's advantages in difficult lighting conditions and when hands are gloved, showing superior performance and better adaptability to challenging settings than other spectral techniques. 
Our main contributions are summarized as follows:

\begin{itemize}[label=$\bullet$]
\item  We introduce the first-of-its-kind benchmark, \sysname, to investigate the potential of thermal imaging for egocentric 3D hand pose estimation.
\item We collected a diverse dataset comprising approximately 96,000 synchronized multi-spectral, multi-view images capturing hand-object and hand-virtual interactions from 28 participants across various environments. This dataset is enriched with 3D hand pose ground truths through an innovative automatic annotation process. 
\item We introduce a new baseline method, termed \emph{TherFormer}, and implement state-of-the-art image-based methods on our dataset for benchmarking. 
\item Based on the \sysname benchmark, we conduct comprehensive experiments and analysis on TherFormer and state-of-the-art methods.
\item We release our dataset, code and models and maintain the benchmark at \url{https://github.com/LawrenceZ22/ThermoHands}.
\end{itemize}

\section{Related Works}
\begin{table*}[!tbp]
    \renewcommand\arraystretch{1.0}
    \setlength\tabcolsep{15pt}
    \centering
    \resizebox{2\columnwidth}{!}{%
    \begin{tabular}{@{}lccccccccc@{}}
    \toprule
    & \multicolumn{4}{c}{Normal office (Main)} & \multicolumn{4}{c}{Other settings} &\multirow{2}{*}{Total} \\
        \cmidrule(r){2-5} \cmidrule(r){6-9} 
    \textbf{Setting}  &  train & val & test & sum & darkness & sun glare & gloves & kitchen \\
    \midrule
      \#frames  & 47,436 & 12,914 & 24,002 & 84,352 & 3,188 & 2,508 & 3,068 & 2,808 & 95,924\\
      \#seqs & 172 & 43 & 86 & 301 &12 & 12 & 12 & 14 & 352\\
      \#subjects & 16 & 4 & 8 & 28 & 1 & 1 & 1 & 2 & -\\ 
    \bottomrule
    \end{tabular}
    }
    \caption{\textbf{Benchmark Dataset Statistics}. The overall duration of our dataset is over 3 hours with $\sim$96K synchronized frame of all types of images collected. }
    \label{tab:stat}
\end{table*}

\subsection{3D Hand Pose Datasets}
Datasets with 3D hand pose annotations are imperative for training and evaluating \emph{ad-hoc} models. Existing datasets, according to their approaches of annotation acquisition, can be summarized as four types in general, \ie, marker-based~\cite{yuan2017bighand2, garcia2018first, taheri2020grab,fan2023arctic}, synthetic~\cite{mueller2017real, hasson2019learning, mueller2019real, mueller2018ganerated, zimmermann2017learning}, manual~\cite{mueller2017real,sridhar2016real} or hybrid~\cite{chao2021dexycb, zimmermann2019freihand, moon2020interhand2, liu2022hoi4d, ohkawa2023assemblyhands}, and automatic~\cite{hampali2020honnotate,brahmbhatt2020contactpose,kwon2021h2o,sener2022assembly101} annotated datasets. Marker-based approaches, using magnetic sensor~\cite{yuan2017bighand2, garcia2018first} or Mocap markers~\cite{taheri2020grab,fan2023arctic}, can alter and induce bias to the hand appearance. Synthetic data~\cite{mueller2017real, hasson2019learning, mueller2018ganerated, zimmermann2017learning,mueller2019real} suffers from the \emph{sim2real} gap in terms of hand motion and texture features. Introducing human annotators to fully~\cite{mueller2017real,sridhar2016real} or partly~\cite{chao2021dexycb, zimmermann2019freihand, moon2020interhand2, liu2022hoi4d, ohkawa2023assemblyhands} annotate 2D/3D keypoints circumvents the issues above, but it either limits the scale of datasets or manifests costly and laborious in practice. Most similar to ours, some datasets adopt fully automatic pipelines to obtain 3D hand pose annotations~\cite{hampali2020honnotate,brahmbhatt2020contactpose,kwon2021h2o}, which leverage pre-trained models (\eg OpenPose~\cite{8765346}) to infer the prior hand information and rely on optimization to fit the MANO hand model~\cite{MANO}. 

Despite the existing progress, previous datasets only provide depth~\cite{yuan2017bighand2, mueller2019real}, RGB images~\cite{mueller2018ganerated, moon2020interhand2, zimmermann2019freihand, ohkawa2023assemblyhands} or both of them~\cite{garcia2018first, taheri2020grab, hasson2019learning,zimmermann2017learning,mueller2017real,sridhar2016real,chao2021dexycb, liu2022hoi4d,hampali2020honnotate,brahmbhatt2020contactpose,kwon2021h2o} as the input spectra, unable to support the study of NIR or thermal image-based 3D hand pose estimation. \sysname fills the gap by providing a moderate amount of multi-spectral image data, from infrared to visual light, paired with depth images. Moreover, we capture bimanual actions from both egocentric and exocentric viewpoints and design hand-object as well as hand-virtual interaction actions to facilitate a wide range of applications. 

\subsection{Image-based 3D Hand Pose Estimation}
As a key computer vision task, 3D hand pose estimation from images is highly demanded by applications like XR~\cite{wu2020hand,sagayam2017hand,marchand2015pose,liang2015ar},  human-robot interaction~\cite{sampieri2022pose,gao2021dynamic,gao2019dual} and imitation learning~\cite{qin2022dexmv,chen2023bi,antotsiou2018task}. Therefore, this field has been extensively explored in previous arts that uses single RGB~\cite{ jiang2023a2j,spurr2020weakly,yang2019disentangling, cheng2021handfoldingnet,kim2021end,lin2021two, moon2020interhand2,li2022interacting,smith2020constraining,wang2020rgb2hands,zhang2021interacting, hasson2019learning,liu2021semi,boukhayma20193d,baek2019pushing, zhang2019end} or 
depth~\cite{ge2018robust,moon2018v2v,xiong2019a2j,zhang2020handaugment,zhu2019hand} image as input. 
These methods can be roughly categorized into two fashions, \ie, model-based and model-free methods. Model-based methods~\cite{li2022interacting,smith2020constraining,wang2020rgb2hands,zhang2021interacting, hasson2019learning,liu2021semi,boukhayma20193d,baek2019pushing, zhang2019end} utilize the prior knowledge of the MANO hand model~\cite{MANO} by estimating its shape and pose parameters, while model-free methods~\cite{jiang2023a2j,spurr2020weakly,yang2019disentangling, cheng2021handfoldingnet,kim2021end,lin2021two, moon2020interhand2,ge2018robust,moon2018v2v,xiong2019a2j,zhang2020handaugment,zhu2019hand} learn the direct regression of 3D hand joints or vertices coordinates. 
Recently there has been growing interest in leveraging the temporal supervision~\cite{hasson2020leveraging,liu2021semi,yang2020seqhand,Kocabas_2020_CVPR,park2022handoccnet} or leveraging sequential images as input~\cite{wen2023hierarchical, Fu_2023_ICCV,cho2023transformer, Cai_2019_ICCV,khaleghi2022multi,choi2021beyond} for 3D hand pose estimation. In this study, we evaluate existing methods and our baseline method in both single image-based and video-based problem settings, respectively.
Apart from the previous approaches, we investigate the potential of thermal imagery for tackling various challenges in 3D hand pose estimation. 
\begin{figure}[!tbp]
    \centering
    \includegraphics[width=.5\textwidth]{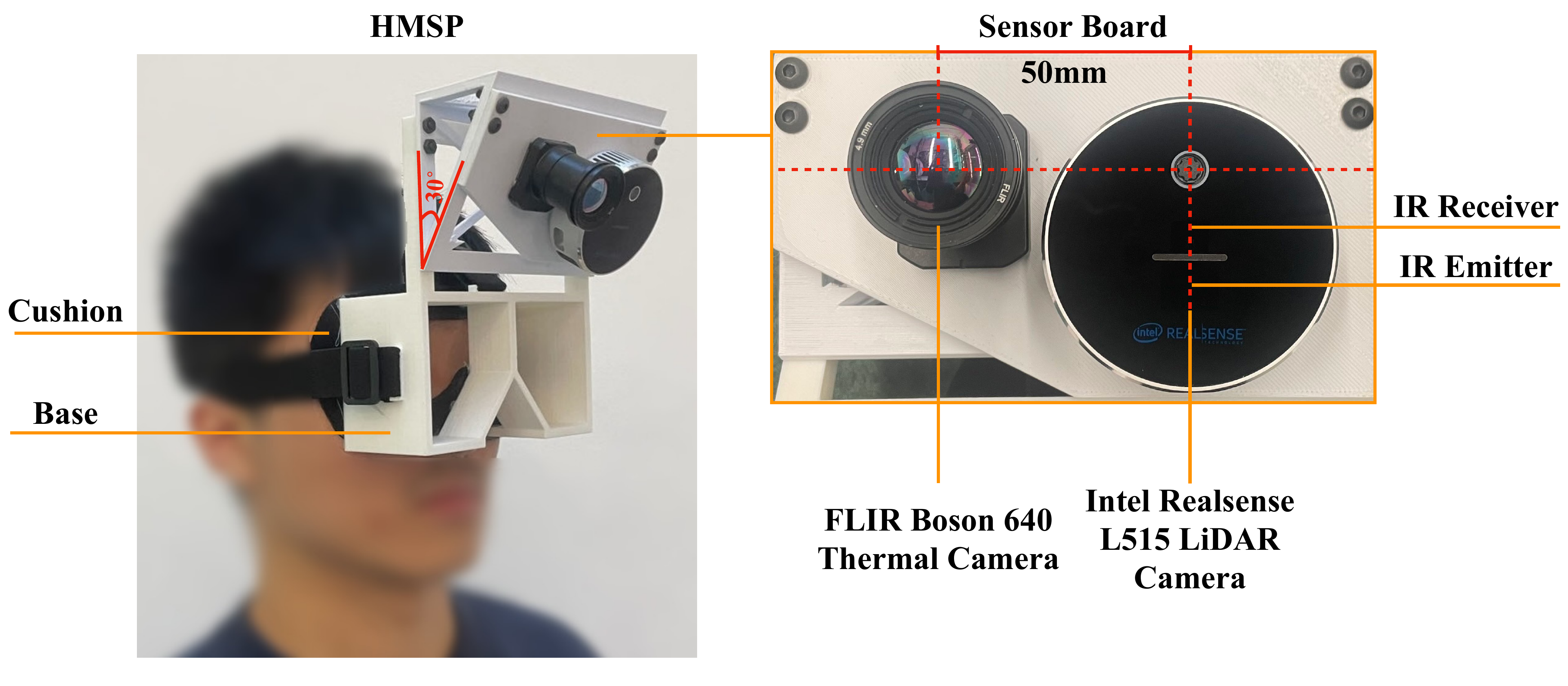}
    \caption{Design of the head-mounted sensor platform and sensor alignment.}
    \label{fig:HMSP}
\end{figure}

\subsection{Thermal Computer Vision}

Thermal cameras achieve imaging by capturing the radiation emitted 
in the LWIR spectrum and deducing the temperature distribution on the surfaces~\cite{lloyd2013thermal}. Leveraging its robustness to variable illumination and unique temperature information, numerous efforts have been made to address various computer vision tasks, including super-resolution~\cite{rivadeneira2023thermal,gupta2021toward,kansal2020multi}, human detection~\cite{10.5555/3451271.3451272,ivavsic2019human,haider2021human}, action recognition~\cite{batchuluun2019action,ding2022individual} and pose estimation~\cite{lupion20243d,chen2020multi,smith2023human}, semantic segmentation~\cite{kutuk2022semantic,kim2021ms,vertens2020heatnet,li2020segmenting}, depth estimation~\cite{Shin_2023_CVPR, lu2021alternative,shin2021self,kim2018multispectral}, visual(-inertial) odometry/SLAM~\cite{rob.21932,9623261,shin2019sparse,saputra2020deeptio}, 3D reconstruction~\cite{liu2023humans,sage20213d,schramm2022combining}, \emph{etc}. In this work, we focus on 3D hand pose estimation, which is an under-exploited task based on thermal images.

\begin{figure*}[!tbp]
    \begin{subfigure}{.24\textwidth}
        \centering
        \includegraphics[width=\linewidth]{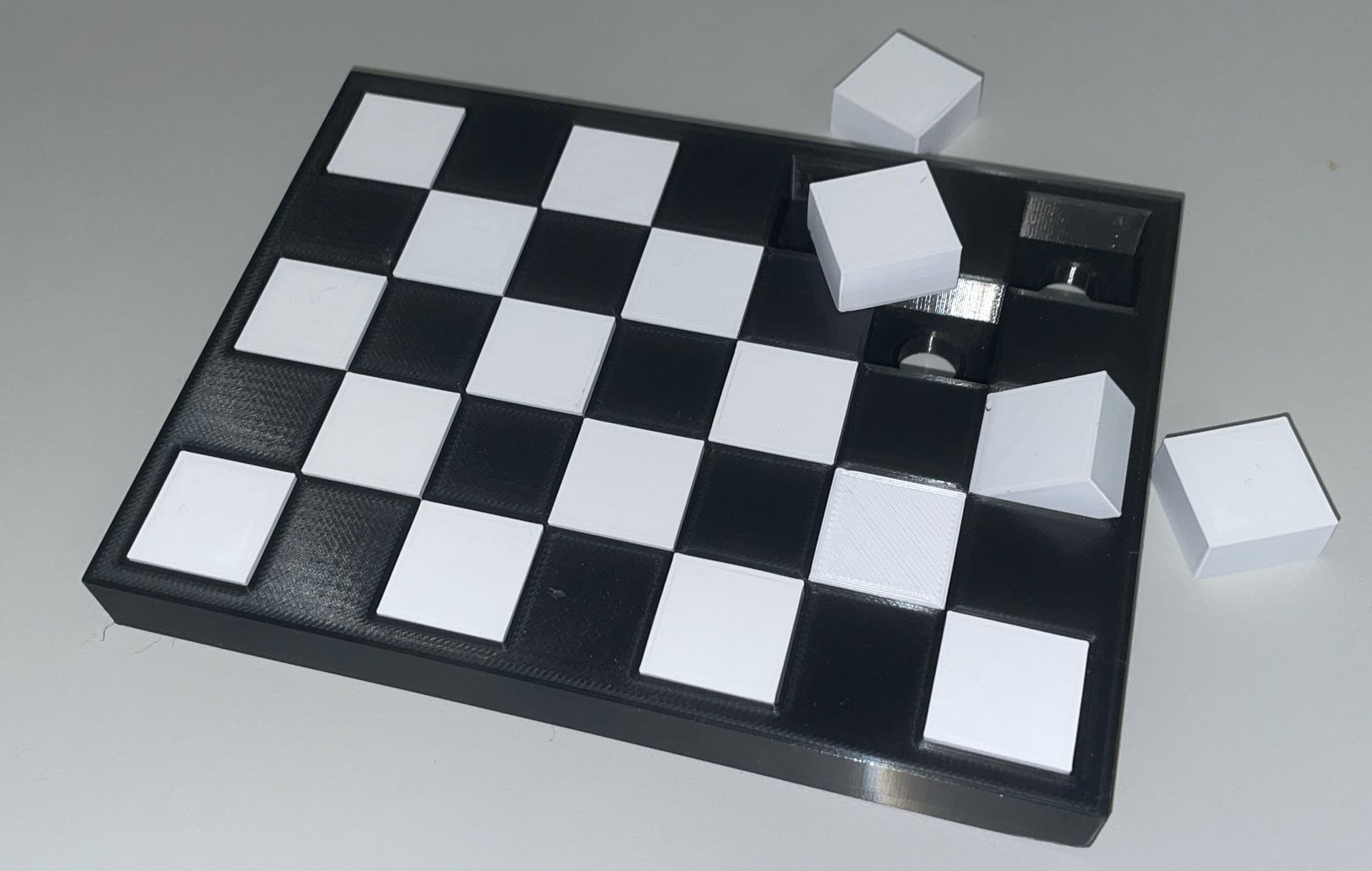}
        \caption{}
        \label{fig:board}
    \end{subfigure}
    \begin{subfigure}{.24\textwidth}
        \centering
        \includegraphics[width=\linewidth]{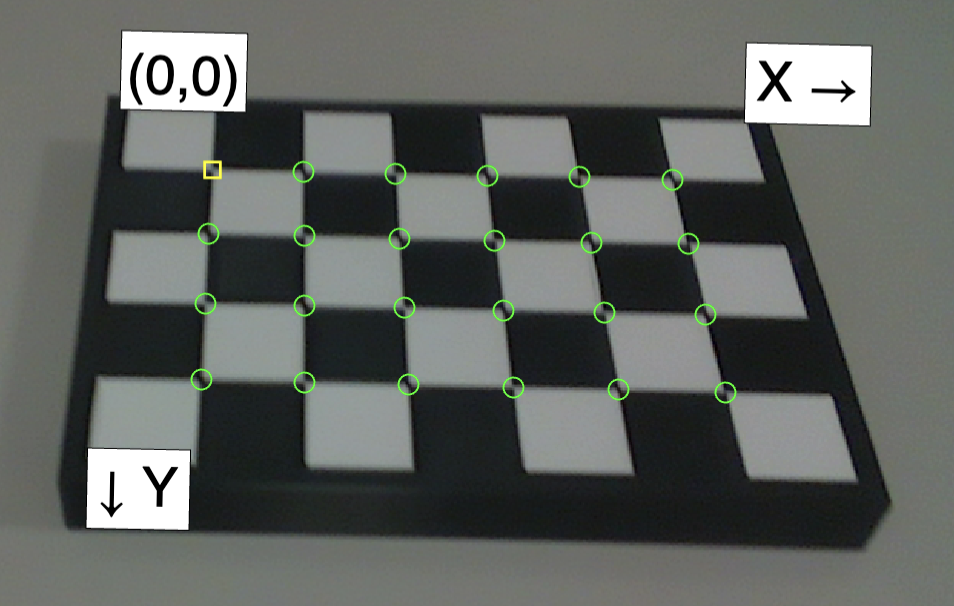}
        \caption{}
        \label{fig:sub1}
    \end{subfigure}\hfill
    \begin{subfigure}{.24\textwidth}
        \centering
        \includegraphics[width=\linewidth]{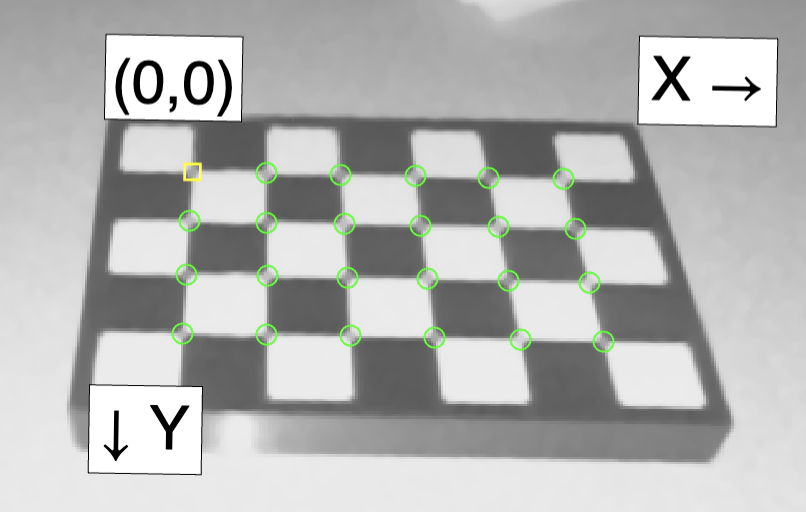}
        \caption{}
        \label{fig:sub2}
    \end{subfigure}\hfill
    \begin{subfigure}{.24\textwidth}
        \centering
        \includegraphics[width=\linewidth]{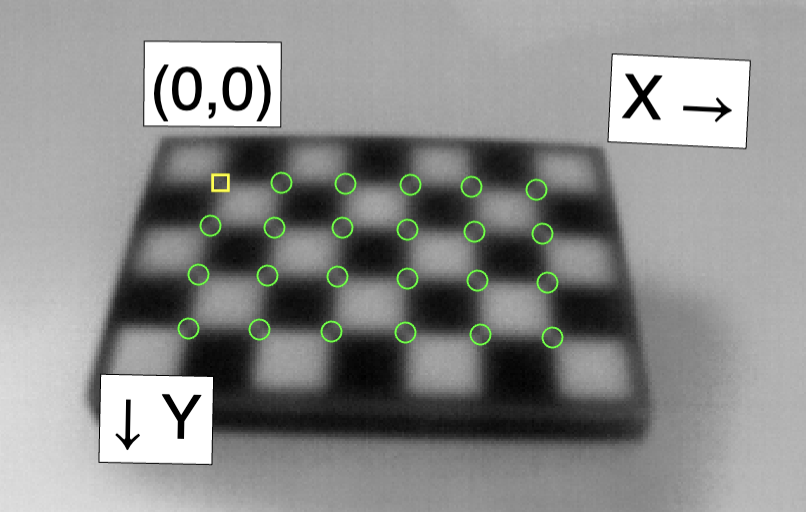}
        \caption{}
        \label{fig:sub3}
    \end{subfigure}
    \caption{\textbf{Thermal calibration chessboard} containing a black base board and multiple removable white cubes (a). By cooling down the base board, it shows similar patterns and allows automatic corner detection in all (b) RGB, (c) NIR and (d) thermal images.}
    \label{fig:thermal_calib_thermal}
\end{figure*}
\section{The ThermoHands Benchmark}

\subsection{Multi-Spectral Hand Pose Dataset}\label{dataset}

\noindent\textbf{Overview.} At the core of our benchmark lies a multi-spectral dataset for 3D hand pose estimation (\cf~\cref{tab:stat}), capturing hand actions performed by 28 subjects of various ethnicities and genders\footnote{The study has received the ethical approval from University of Edinburgh,
and participant consent forms were signed before the collection.}. As shown in~\cref{fig:capture}, we develop a customized head-mounted sensor platform (HMSP) and an exocentric platform to record multi-view data. During the capture, our participants are asked to perform pre-defined hand-object and hand-virtual interaction actions within the playground above the table. The main part is captured in the normal office scenario. To facilitate the evaluation under different settings, four auxiliary parts are recorded i) under the darkness, ii) under the sun glare, iii) with gloves on hand, and iv) in the kitchen environment with different actions, respectively.   

\noindent\textbf{Sensor Platforms.}
Apart from the traditional approach of mounting cameras on helmets \cite{kwon2021h2o}, the design of the HMSP (\cf~\cref{fig:HMSP}) focuses on simulating an actual Mixed Reality (MR) device and reducing extra weight to allow participants to perform freely. The HMSP consists of three major components: a cushion for comfort, a base component that provides a 30-degree downward tilt, and a sensor board that carries two cameras - an Intel RealSense L515 LiDAR camera~\cite{IntelRSL515} that streams egocentric RGB, depth, and NIR images, and a Teledyne FLIR Boson 640 long-wave infrared (LWIR) camera~\cite{FLIRBoson2024} that receives the LWIR to obtain the thermal images.
An extra exocentric platform equipped with an Intel RealSense D455~\cite{IntelRealSense2023D455}
is leveraged to support multi-view annotation (\cf \cref{sec:annotation}) as well as provide the RGB-D image data from the third-person viewpoint. 
As exhibited in~\cref{fig:capture}, we place the two depth sensors outside each other's field of view (FoV) to minimize interference caused by their NIR emitters~\cite{IntelRSL515, IntelRealSense2023D455}. 


\noindent\textbf{Synchronization.} We use a single PC to simultaneously gather data streams from two sensor platforms, ensuring the synchronization of their timestamps. After collection, we synchronize six types of images, each with distinct frame rates, \wrt the timestamps of thermal images (8.5fps), thereby generating well-aligned multi-spectral, multi-view data samples as our released data.

\noindent\textbf{Egocentric Calibration.}
For accuracy, factory-calibrated intrinsic parameters are used for the Intel RealSense D455 and L515 cameras~\cite{IntelRSL515, IntelRealSense2023D455}. 
As seen in~\cref{fig:HMSP}, although the thermal camera's center is aligned with the RGB camera's center at the same height on the 3D-printed sensor board, we still calibrate the extrinsic parameters between these two egocentric cameras to ensure enhanced alignment accuracy. To accommodate the unique imaging mechanism of thermal camera, we self-design a modular calibration chessboard as shown in~\cref{fig:thermal_calib_thermal}. Before calibration, we cool down the black base board while keeping the white cubes at room temperature to create a visible chessboard pattern in all three spectra, which can be seen in~\cref{fig:thermal_calib_thermal}. In this way, we can simultaneously calibrate the intrinsic parameter of the thermal camera and its extrinsic parameter \wrt the D455 RGB-D camera. 

\noindent\textbf{Cross-View Calibration.} To calibrate the cameras between the egocentric and exocentric viewpoints, we place 11 IR reflective spheres at random heights within the playground, ensuring they are visible from both viewpoints. Initially, we plan to detect these spheres automatically from the NIR images and track them with the Kalman Filter~\cite{wan2000unscented}. However, we find this approach leads to many false positive detections, which severely affect the tracking accuracy. To ensure the calibration accuracy, we only annotate the sphere markers manually from two viewpoints in the first frame. The 3D positions of markers can be computed using the egocentric depth image and the cross-view transformation can be computed by solving the PnP~\cite{FischlerBolles1981}. For the subsequent frames, we can calculate the transformation by only tracking the pose of the egocentric camera as the exocentric platform keeps stationary during collection. We leverage the state-of-the-art odometry method KISS-ICP~\cite{vizzo2023kiss} and track the motion of the egocentric camera with the point clouds converted from depth images. Specifically, we set the  \texttt{initial\_threshold} and \texttt{min\_motion\_th} parameters of KISS-ICP as 0.0001 to make it capable of catching subtle motion. The input point cloud range is set as [0.2, 2.0] meters while the \texttt{max\_points\_per\_voxel} is 30.



\noindent\textbf{Dataset Statistics.}
As seen in~\cref{tab:stat}, our dataset consists of approximately 96K synchronized multi-spectral multi-view frames (\cf~\cref{fig:capture}) and 352 independent sequences in total. The main part is collected under the normal office scenario, where each participant\footnote{Due to their limited time, 7 participants only perform the hand-object actions.} performs 7 hand-object interaction actions: \emph{cut paper, fold paper, pour water, read book, staple paper, write with pen, write with pencil}, and 5 hand-virtual interaction actions: \emph{pinch and drag, pinch and hold, swipe, tap, touch}, with two hands. This main part is divided into the training, validation and testing splits by subjects with a ratio of 4:1:2. We also collect four auxiliary testing sets by asking one subject to perform the aforementioned 12 actions in the darkness, sun glare and gloves settings individually, and two subjects to perform 7 scenarios-specific interaction actions: \emph{cut, spray, stir, wash hands, wash mug, wash plate, wipe} in the kitchen. 


\begin{figure*}[!tbp]
    \centering
    \includegraphics[width=\textwidth]{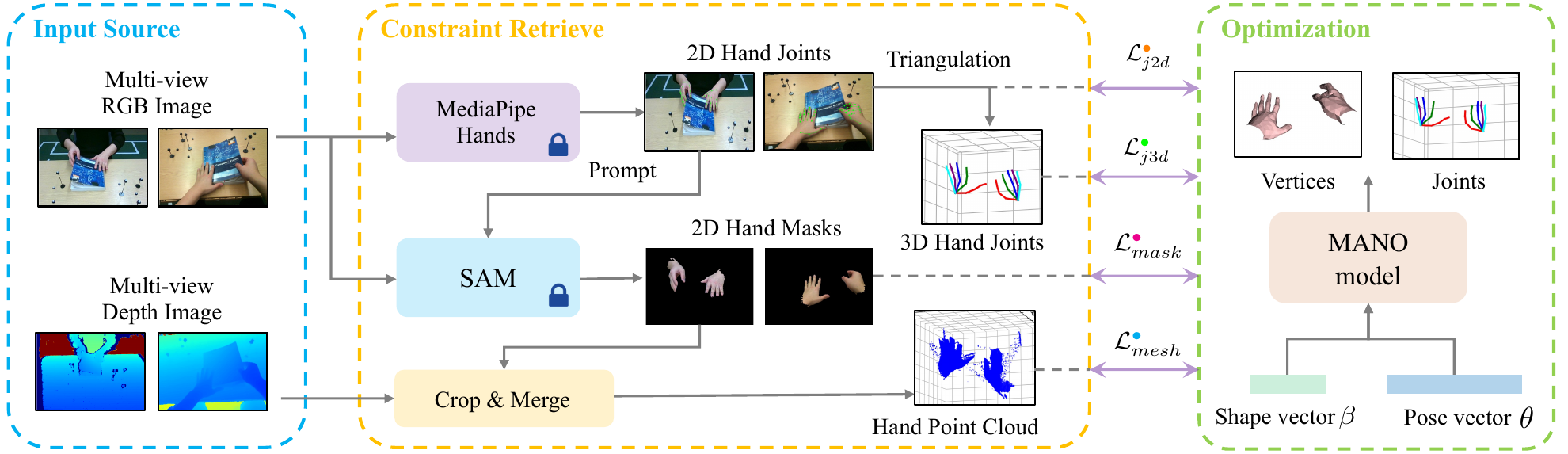}
    \caption{{\textbf{Automatic annotation pipeline} of 3D hand pose. We utilize the multi-view RGB and depth images as the input source and retrieve constraint information with off-the-shelf MediaPipe Hands~\cite{mediapipe_hands_2020} and SAM~\cite{kirillov2023segment}. Various error terms are formulated to optimize the MANO parameters.}}
    \label{fig:anno}
\end{figure*}
\section{Hand Pose Annotation}\label{sec:annotation}
To avoid employing tedious human efforts for annotation, we implement a fully automatic annotation pipeline, similar to the approaches in~\cite{hampali2020honnotate,brahmbhatt2020contactpose,kwon2021h2o}, to obtain the 3D hand pose ground truth for our dataset. 
In particular, we use the MANO statistical hand model~\cite{MANO} to represent 3D hand pose. The MANO model parameterizes the hand mesh vertices $\mathcal{V} (\beta, \theta)$ 
into two low-dimensional embeddings, \ie, the shape parameters $\beta\in\mathbb{R}^{10}$ and the pose parameters $\theta\in\mathbb{R}^{51}$, consisting of 45 parameters accounting for 15 hand joint angles (3 DoF for each) plus the rest for global rotation and translation. Following the MANO PyTorch version in~\cite{hasson19_obman}, we denote the $N_{\mathcal{J}} = 21$ hand joints mapped from the hand parameters as $\mathcal{J}(\beta, \theta)$.
The MANO fitting is performed by minimizing the following optimization objective per frame for each hand: 
\begin{equation}
\begin{aligned}
       \theta^* & = \argmin_{\theta} \lambda_{j2d} \mathcal{L}_{j2d}^{\textcolor{orange}{\bullet}}  + \lambda_{mask} \mathcal{L}_{mask}^{\textcolor{magenta}{\bullet}} +  \lambda_{j3d} \mathcal{L}_{j3d}^{\textcolor{green}{\bullet}} \\ 
       & + \lambda_{mesh} \mathcal{L}_{mesh}^{\textcolor{cyan}{\bullet}} + \lambda_{reg} \mathcal{L}_{reg}^{\textcolor{red}{\bullet}}
\end{aligned}
\label{eq:overall}
\end{equation}
where $\lambda_{j2d}, \lambda_{mask}, \lambda_{j3d}, \lambda_{mesh}, \lambda_{reg}$ are used to balance the weight of different errors. The diagram illustrating our annotation process is shown in~\cref{fig:anno}. 

\noindent\textbf{Initialization.} For each sequence, we optimize the shape parameter $\beta$\footnote{For simplicity, we omit $\beta$ in~\cref{eq:overall} and subsequent equations.} 
together with $\theta$ only at the first frame using \cref{eq:overall} until convergence and fix its values for the subsequent frames. We initialize the pose parameter $\theta$ for each frame using the optimization result from the last frame. This helps to accelerate the convergence as well as keep the temporal consistency. 

\noindent\textbf{2D Joint Error $\mathcal{L}_{j2d}^{\textcolor{orange}{\bullet}}$.} Given RGB images from $N_C$ viewpoints, we infer 2D hand joints $\mathcal{J}^{2D}$ with MediaPipe Hands~\cite{mediapipe_hands_2020} and define the 2D joint error as:
\begin{equation}
    \mathcal{L}_{j2d}^{\textcolor{orange}{\bullet}} = \sum_{c=1}^{N_C}\alpha_{c}
    \sum_{i=1}^{N_\mathcal{J}}||\mathcal{J}^{2D}_{c,i} - \pi_{c} (\mathcal{J}(\theta)_{i})||
\end{equation}
where $\pi_c (\cdot)$ returns the 2D projection location for 3D position in the $c$-th camera viewpoint, and $\alpha_c$ is hyperparameter used to weigh different viewpoints. 

\begin{figure*}[tb!]
  \centering
  \includegraphics[width=\textwidth]{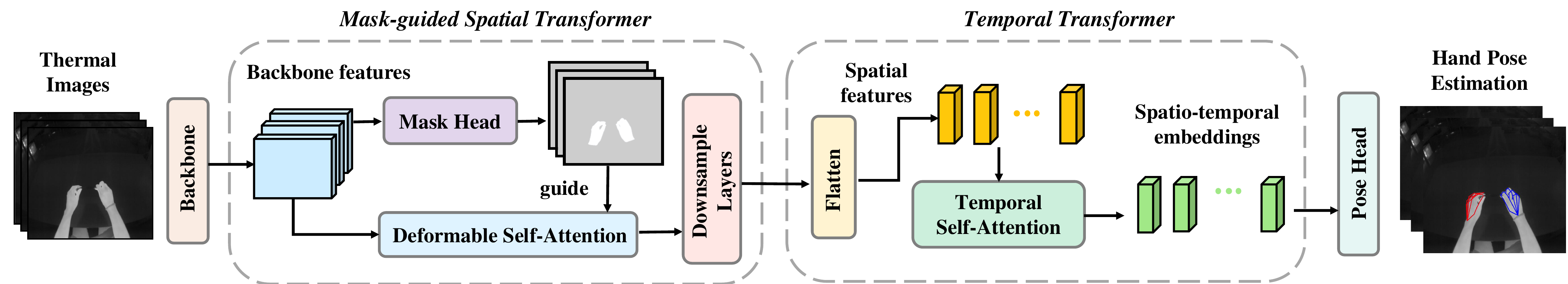}
  \caption{\textbf{Overall Framework of TherFormer.} Backbone features are input to the mask-guided spatial transformer and temporal transformer to enhance the spatial representation and temporal interaction. Spatio-temporal embeddings are fed into the pose head to regress the 3D hand pose. }
  \label{fig:network}
\end{figure*}

\noindent\textbf{2D Mask Error $\mathcal{L}_{mask}^{\textcolor{magenta}{\bullet}}$.} To generate the high-quality 2D hand mask, we prompt the prevalent Segment Anything Model~\cite{kirillov2023segment} with the 2D hand joints $\mathcal{J}^{2D}$ and the bounding box derived from it. We penalize the distance between the hand mesh vertices $\mathcal{V}(\theta)$ and the binary hand mask $\mathcal{M}_{c}$ as:
\begin{equation}
    \mathcal{L}_{mask}^{\textcolor{magenta}{\bullet}} = \sum_{c=1}^{N_C}\alpha_{c} \sum_{i=1}^{N_\mathcal{V}} \min_{j}||\mathcal{M}_{c,j}-\pi_c(\mathcal{V}(\theta)_i)||
\end{equation}
where $\mathcal{M}_{c,j}$ is the coordinate of $j$-th non-zero pixel in the mask $\mathcal{M}_c$. 

\noindent\textbf{3D Joint Error $\mathcal{L}_{j3d}^{\textcolor{green}{\bullet}}$.} We triangulate the 2D joints from multiple views to lift them to 3D joints $\mathcal{J}^{3D}$ and measure their difference to $\mathcal{J}(\theta)$, which is written as:
\begin{equation}
    \mathcal{L}_{j3d}^{\textcolor{green}{\bullet}} = \sum_{i=1}^{N_\mathcal{J}}||\mathcal{J}^{3D}_{i} - \mathcal{J}(\theta)_{i}||
\end{equation}

\noindent\textbf{3D Mesh Error $\mathcal{L}_{mesh}^{\textcolor{cyan}{\bullet}}$.} To better supervise the hand mesh and fasten the optimization, we generate the 3D hand pose cloud $\mathcal{P}$ by cropping the depth image using the 2D hand mask and merging all views together. The 3D mesh error term compensates for the distance between the hand mesh and point cloud:
\begin{equation}
    \mathcal{L}_{mesh}^{\textcolor{cyan}{\bullet}} = \sum_{i=1}^{N_\mathcal{V}} \min_j||\mathcal{P}_{j}-\mathcal{V}(\theta)_i||
\end{equation}

\noindent\textbf{Regularization $\mathcal{L}_{reg}^{\textcolor{red}{\bullet}}$.} To alleviate irregular hand articulation, we constrain the joint angles to pre-defined lower and upper boundaries $\underline{\theta}$ and $\overline{\theta}$:
\begin{equation}
    \mathcal{L}_{reg}^{\textcolor{red}{\bullet}} =  \sum_{i=1}^{45}(\max(\underline{\theta_i}-\theta_i, 0) +
     \max(\theta_i - \overline{\theta_i}, 0)
     ) 
\end{equation}
Note that we also impose regularization to the shape parameter $\beta$ during initialization. For more implementation details of our dataset annotation process, please refer to~\cref{annotation}.

\section{TherFormer: A Baseline Method}\label{sec:network}
For our benchmark, we setup a baseline method, dubbed TherFormer, for thermal image-based 3D hand pose estimation. As exhibited in~\cref{fig:network}, TherFormer features in its two consecutive transformer modules to model the spatio-temporal relationship of hand joints while being computationally efficient. Note that TherFormer is also capable of processing other spectral images due to their format consistency with thermal images. However, our primary objective is to establish a baseline method for future research, rather than focusing exclusively on methods for thermal images.

\noindent\textbf{Problem Definition.} We consider two problem settings: single image-based and video-based egocentric 3D hand pose estimation for our benchmark. In the former setting, we aim to estimate the 3D joint positions $\mathcal{J}_t$ for two hands given the single thermal image $\mathcal{I}_t$ captured for the $t$-th frame. For the video-based one, our input is a sequence of thermal images $\mathcal{S} = \{\mathcal{I}_i\}_{i=1}^{T}$ and we estimate the per-frame 3D hand joint positions $\mathcal{J}_i$ together. 
Without losing the generality, here we illustrate our network architecture for the video-based setting, \ie, TherFormer-V. In practice, our network can be flexibly adapted to the single image-based version (\ie, TherFormer-S) by setting $T=1$.



\begin{table*}[tb!]\small
    \renewcommand\arraystretch{1}
    \setlength\tabcolsep{5pt}
    \centering
    \resizebox{2\columnwidth}{!}{%
    \begin{tabular}{@{}lccccccccccc@{}}
    \toprule
        & \multicolumn{2}{c}{Ego-view optimization} & \multicolumn{4}{c}{Multi-view optimization} \\
        \cmidrule(r){2-3}   \cmidrule(r){4-7} 
        \textbf{Errors} 
        & \multirow{1}{*}{$\mathcal{L}_{mask}+ \mathcal{L}_{j2d}$} 
        & \multirow{1}{*}{\makecell[c]{$\mathcal{L}_{mask}+ \mathcal{L}_{j2d} +  \mathcal{L}_{mesh} 
         $}} 
        & \multirow{1}{*}{$\mathcal{L}_{mask}$} 
        & \multirow{1}{*}{$\mathcal{L}_{mask}+ \mathcal{L}_{j2d}$} 
        & \multirow{1}{*}{\makecell[c]{$\mathcal{L}_{mask}+ \mathcal{L}_{j2d} +  \mathcal{L}_{mesh}$}} 
        & \multirow{1}{*}{\makecell[c]{$\mathcal{L}_{mask}+ \mathcal{L}_{j2d} +  \mathcal{L}_{mesh} + \mathcal{L}_{j3d}$}} 
         \\
    \midrule 
    \textbf{mean (std)} & 37.29 ($\pm$ 18.02) & 7.03 ($\pm$ 2.57) 
    & 8.13 ($\pm$ 0.57) & 1.29 ($\pm$ 0.43) & 1.28 ($\pm$ 0.43) & 1.01 ($\pm$ 0.34) \\
    \bottomrule
    \end{tabular}
    }
     \caption{Evaluation of annotation results. The average 3D joint errors across all frames are reported (in cm). $\mathcal{L}_{reg}$ is used for all to mitigate irregular hand poses.}
    \label{tab:annotation}
\end{table*}

\begin{figure*}[tbp!]
    \centering
    \includegraphics[width=\textwidth]{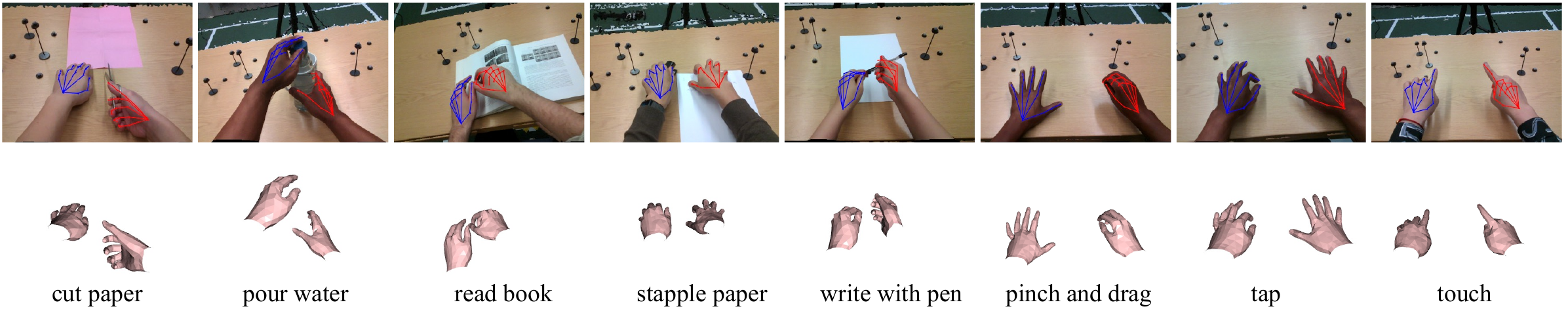}
    \caption{{Examples of 3D hand pose annotations. Top row: left ({\color{blue}{blue}}) and right ({\color{red}{red}}) hand 3D joints projected onto egocentric RGB images. Bottom row: visualization of hand mesh annotation.}}
    \label{fig:mesh}
\end{figure*}

\noindent\textbf{Mask-guided Spatial Transformer.} Human hands are highly articulated objects that can adopt a wide variety of poses, often against complex backgrounds. We propose a mask-guided spatial transformer module to accurately identify and focus on the intricacies of hand poses during spatial feature interaction. Given backbone features, we first utilize a mask head to estimate the binary hand mask in the thermal image. Then, we leverage the deformable self-attention~\cite{zhu2020deformable} to refine the hand spatial features under the guidance of the estimated hand mask. Specifically, we only take feature elements whose spatial locations are within the hand area as queries and sample keys from only the hand area and its surrounding locations. In this way, we not only reduce the computation waste on the irrelevant region but also increase the robustness to background clutter. Lastly, we reduce the spatial dimensions of the spatial features with a series of convolutions layers for efficiency. 

\noindent\textbf{Temporal Transformer.} Temporal information is crucial for 3D hand pose estimation when coping with occlusion and solving ambiguities. To model temporal relationships, we first flatten the spatial features into 1D feature vectors and then employed the temporal self-attention~\cite{vaswani2017attention} to explicitly attend to the feature vector of every frame. The output is frame-wise spatio-temporal feature embeddings. Note that the temporal self-attention degrades to an MLP for single-image based setting. 

In the pose head, we use the MLP to project the embeddings to the output space and obtain the per-frame 3D joint $\mathcal{J}_i$. We leverage the binary cross-entropy loss to supervise the hand segmentation with the mask ground truth rendered from the annotated hand mesh (\cf~\cref{sec:annotation}). For 3D hand joint positions, we measure the \emph{L1} distance of its 2D projection and depth to that of the ground truth separately. 




\begin{table*}[tb!]\small
    \renewcommand\arraystretch{1}
    \setlength\tabcolsep{16pt}
    \centering
    \resizebox{2\columnwidth}{!}{%
    \begin{tabular}{@{}c|lcccccc@{}}
    \toprule
       & Method & Input & MEPE (mm) $\downarrow$ & AUC $\uparrow$ & MEPE-RA (mm) $\downarrow$ & AUC-RA $\uparrow$ & fps $\uparrow$\\
    \midrule 
    (a) & HaMeR*~\cite{pavlakos2024reconstructing} & Single & - & - & 20.88 & 0.598 & 118 \\
   (b) & A2J-Transformer~\cite{jiang2023a2j} & Single & 51.68 &  0.474 & 20.76 &	0.603 & 34\\
   (c) & HTT~\cite{wen2023hierarchical} & Single & 49.09 & 0.489 & 20.69 & 0.599 & \textbf{211}\\
   (d) & TherFormer-S & Single & \textbf{44.64} &  \textbf{0.539} & \textbf{18.34} & \textbf{0.643} & 136 \\
    \midrule
   (e) &  (c) w/o spatial transformer & Single & 48.79 &  0.491 & 20.15 & 0.609 & 174\\
    (f) & (c) w/o mask guidance & Single  & 48.83 & 0.494 & 18.89 & 0.625 & 141\\
    \midrule
   (g) & HTT~\cite{wen2023hierarchical} & Sequence & 47.07 & 0.512 & 17.49 & 0.659 & \textbf{129}  \\
    (h) & TherFormer-V & Sequence & \textbf{43.36} & \textbf{0.549} & \textbf{17.36} & \textbf{0.661} & 52\\
    \bottomrule
    \end{tabular}
    }
    \caption{Comparison between TherFormer and state of the arts on thermal camera-based 3D hand pose estimation. The fps indicates the number of inference steps that models can run per second. HaMeR* estimates 3D hand pose in a root-aligned manner by default; thus, we can only evaluate MEPE-RA and AUC-RA metrics.}
    \label{tab:results}
\end{table*}

\section{Experiment}\label{experiment}
\subsection{Evaluation of the Annotation Method}
As the first step, we validate the accuracy of our 3D hand pose annotation (\cf~\cref{sec:annotation}) and analyze the impact on optimization results. For evaluation, we manually annotate two random sequences from our main dataset, with a total of over 600 frames. To that end, we first annotate the 2D joint locations on RGB images from two viewpoints and obtain the 3D joint positions by triangulation. We calculate the average 3D joint errors across all frames to measure the accuracy.  

As shown in~\cref{tab:annotation}, our annotation method achieves an average joint error of nearly 1cm, comparable to the results of~\cite{kwon2021h2o, hampali2020honnotate, zimmermann2019freihand}. The multi-view setting shows remarkably better precision than the ego-view only optimization, demonstrating the necessity of multi-camera capture. We also observe that only combining $\mathcal{L}_{mask}$ and $\mathcal{L}_{j2d}$ can already provide a plausible accuracy since they fit the projection of the 3D hand pose to two heterogeneous views. $\mathcal{L}_{mesh}$, though it hardly improves the joint accuracy, can result in more natural hand mesh. Adding $\mathcal{L}_{j3d}$ further refines the joints as it induces the explicit constraint to their positions. We showcase some annotation examples in~\cref{fig:mesh}. As can be seen, both hand joint and mesh can be accurately annotated across different actions despite the presence of occlusion and the variance in subjects' hand color and shape. Please see our \href{https://thermohands.github.io/}{\textbf{project page}} for more visualization of annotation.

\begin{figure*}[tbp!]
    \centering
    \includegraphics[width=0.8\textwidth]{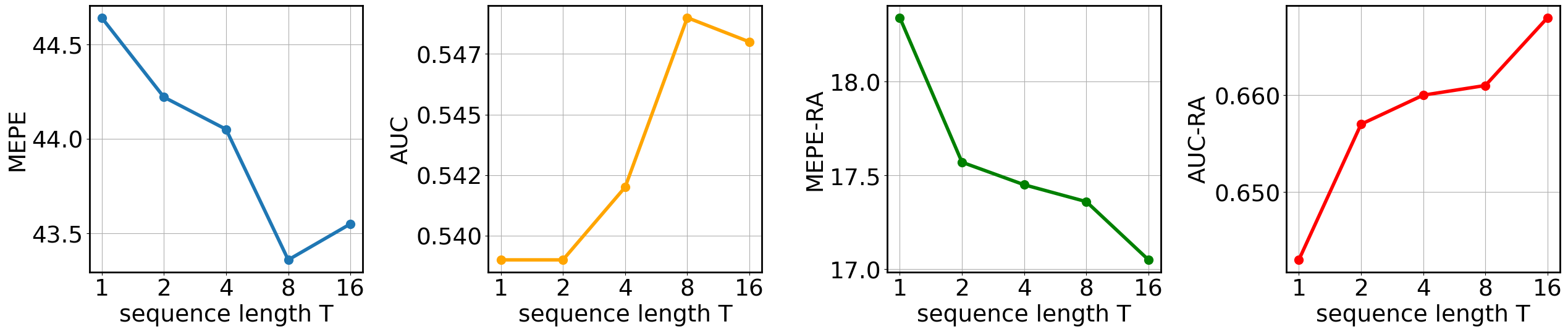}
    \caption{Impact of the temporal sequence length $T$ to TherFormer. The plots show all four metrics  MEPE (mm) $\downarrow$, AUC $\uparrow$, MEPE-RA (mm) $\downarrow$, and AUC-RA $\uparrow$ against 5 sequence length settings, \ie, $T\in$\{1, 2, 4, 8, 16\}.}
    \label{fig:length}
\end{figure*}

\subsection{Experiment Setup}\label{setup}

\noindent\textbf{Dataset Preparation.} We utilize our own dataset for experiments as it uniquely contains egocentric images from multiple spectra, essentially for our benchmark experiments. We annotate the main part of our dataset (\cf~\cref{tab:stat}) automatically following~\cref{sec:annotation}, of which the training and validation sets serve as the foundation for the training of all network models. Our automatic annotation pipeline (\cf\cref{sec:annotation}) becomes infeasible under challenging scenarios, \ie, gloves, darkness and sun glare,  since hands appear corrupted in either RGB or depth images. To facilitate the quantitative evaluation under challenging settings, we manually annotate the ground truth for a few sequences collected in the glove and sun glare scenarios (\cf~\cref{tab:stat}). Specifically, we first annotate the 2D keypoints from two viewpoints and then use triangulation to obtain their 3D positions.

\noindent\textbf{Method and Implementation.} To provide sufficient baselines for follow-up works, we selected three state-of-the-art methods in 3D hand pose estimation: HTT~\cite{wen2023hierarchical}, A2J-Transformer~\cite{jiang2023a2j} and HaMeR~\cite{pavlakos2024reconstructing}, and reproduced them on our dataset for experiments. 
HTT~\cite{wen2023hierarchical} is a video-based method thus enabling the evaluation in both two problem settings while A2J-Transformer~\cite{jiang2023a2j} only works under the single image-based setting. HaMeR~\cite{pavlakos2024reconstructing} is a model-based method that takes a single image as input and reconstructs a MANO-based hand mesh. It estimates each hand individually from a cropped image patch centered on the hand, removing the need for explicit localization. For comparison, we extract 3D keypoints from its predicted mesh as the estimated pose. We use the same sequence length, \ie, $T=8$, for HTT and TherFormer-V baselines. We exclude the additional action block used by HTT~\cite{wen2023hierarchical}, focusing solely on hand pose estimation.
We adjusted the anchor initialization phase of A2J-Transformer~\cite{jiang2023a2j} to better accommodate our dataset, without altering the density of its anchors. All trained models are tested on a single NVIDIA RTX 4090 GPU to fairly compare their inference speed. 

\noindent\textbf{Evaluation Metrics.}  
Following HTT~\cite{wen2023hierarchical}, we evaluate the accuracy of 3D hand pose estimation with two metrics: \emph{Percentage of Correct Keypoints} (PCK) and \emph{Mean End-Point Error} (MEPE)~\cite{zimmermann2017learning}, in both camera space and root-aligned (RA) space. For the RA space, we align the estimated wrist with its groundtruth position before measuring the joint errors. In terms of PCK, we report the corresponding \emph{Area Under the Curve} (AUC) over the 0-80mm/50mm error thresholds for the camera/RA space.





\subsection{Thermal Image-based Results}

\noindent\textbf{Main Results.}
To assess TherFormer's performance in thermal image-based 3D hand pose estimation, we compare it against state-of-the-art methods~\cite{wen2023hierarchical,jiang2023a2j} on the main testing set, as shown in~\cref{tab:results}. TherFormer-S outforms three competing methods under the single image-based setting, while TherFormer-V surpasses the counterpart HTT~\cite{wen2023hierarchical} given the same sequential images as input. Such an improvement mainly stems from our mask-guided spatial attention design that can adaptively encode the spatial interaction among hand joints with the guidance of the hand mask (\cf~\cref{sec:network}). A performance gap can be observed between single image-based and video-based settings for both HTT~\cite{wen2023hierarchical} and TherFormer. We credit this to their usage of temporal information that helps to tackle the occlusion cases and solve ambiguities.  
Thanks to our lightweight network design, TherFormer is highly efficient to run with fps of 136 and 52 for single and sequence input respectively, ensuring its real-time application to resource-constrained devices. Moreover, TherFormer-V also improves the performance over HTT~\cite{wen2023hierarchical} for other spectra (\cf~\cref{tab:spectra}), proving its adaptability to different inputs.

\begin{figure*}[tbp!]
    \centering
    \includegraphics[width=\textwidth]{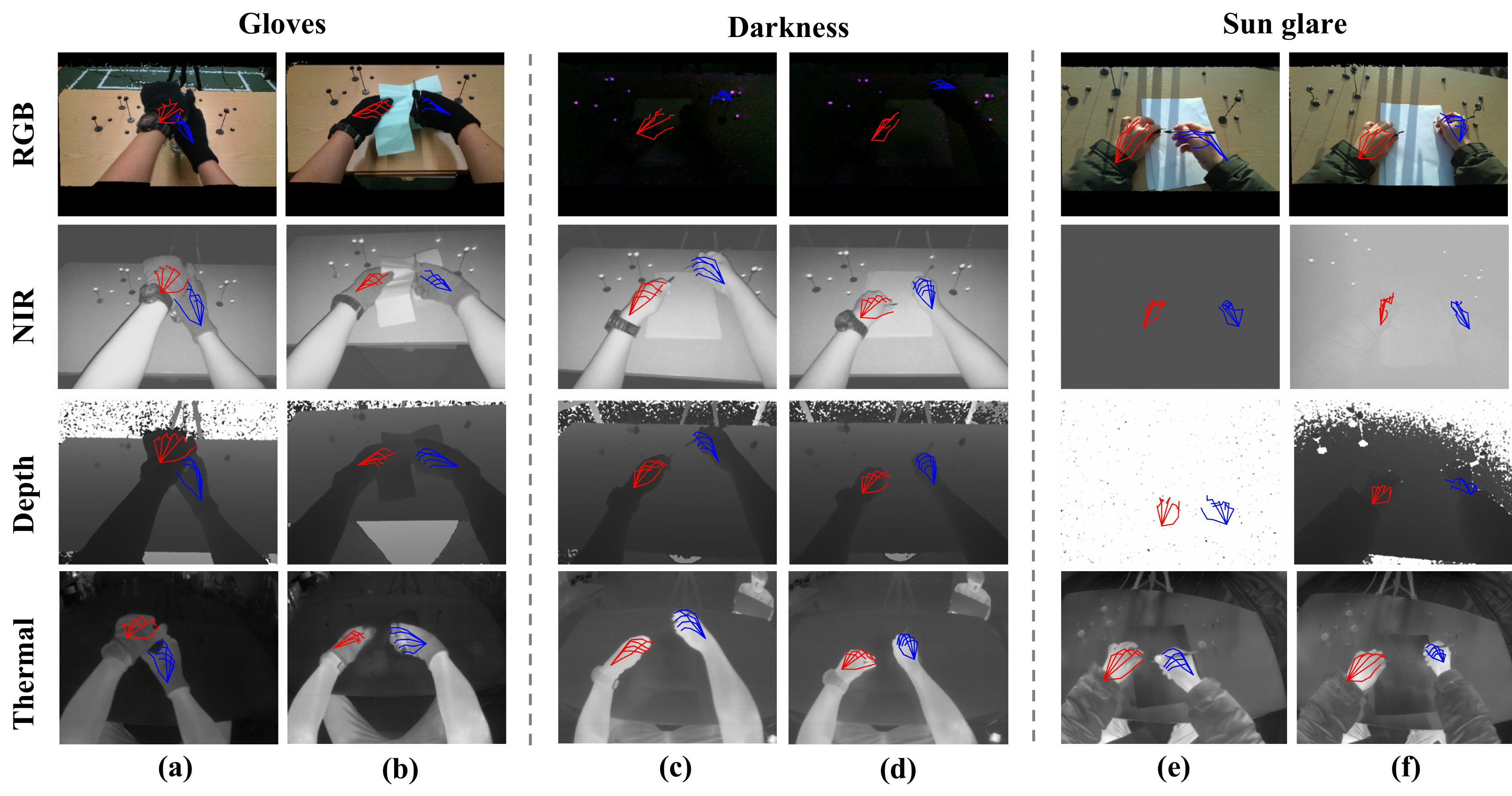}
    \caption{{Examples of results with various spectra under three challenging settings. For visualization, we show the projection of the estimated left ({\color{red}{red}}) and right ({\color{blue}{blue}}) 3D hand joints on images. }}
    \label{fig:challenge}
\end{figure*}

\begin{table}[tb!]\small
    \renewcommand\arraystretch{1}
    \setlength\tabcolsep{10pt}
    \centering
    \resizebox{\columnwidth}{!}{%
    \begin{tabular}{@{}l|cc|cc@{}}
    \toprule
        & \multicolumn{2}{c|}{TherFormer-V (glove)} & \multicolumn{2}{c}{TherFormer-V (sun glare)} \\
    \midrule 
     Spectrum & MEPE-RA (mm) $\downarrow$ & AUC $\uparrow$ & MEPE-RA (mm) $\downarrow$ & AUC $\uparrow$\\
    \midrule 
    RGB & 51.94 & 0.141 & 38.24 & 0.252 \\
    Depth & 45.96 & 0.206 & 42.27 & 0.254 \\
    NIR & 39.83 & 0.282	& 90.84	& 0.093\\
    Thermal & \textbf{39.23} & \textbf{0.302} & \textbf{32.56} & \textbf{0.363}\\
    \bottomrule
    \end{tabular}
    }
    \caption{Comparison between different spectra under two challenging conditions, \ie, glove and sun glare.  }
    \label{tab:challenge}
\end{table}

\begin{table*}[tb!]\small
    \renewcommand\arraystretch{1}
    \setlength\tabcolsep{14pt}
    \centering
    \resizebox{2\columnwidth}{!}{%
    \begin{tabular}{@{}l|cc|cc|cc|cc@{}}
    \toprule
        & \multicolumn{2}{c|}{HTT (Sequence)} & \multicolumn{2}{c|}{TherFormer-V} & \multicolumn{2}{c|}{TherFormer-S} & \multicolumn{2}{c}{Best} \\
    \midrule 
     Spectrum & MEPE (mm) $\downarrow$ & AUC $\uparrow$ & MEPE (mm) $\downarrow$ & AUC $\uparrow$ &  MEPE (mm) $\downarrow$ & AUC $\uparrow$ & MEPE (mm) $\downarrow$ & AUC $\uparrow$\\
    \midrule 
    RGB & 43.30 & 0.542   
    & 43.50 & 0.542  & 44.61 & 0.529 & 43.30 & 0.542 \\
    Depth & 41.62 & 0.559  & \textbf{39.70} & \textbf{0.581} & \textbf{39.84} & \textbf{0.579} & \textbf{39.70} & \textbf{0.581}\\
    NIR & \textbf{41.57} & \textbf{0.562} & 40.79 &  0.575 & 40.98 &  0.575 & 40.79 & 0.575\\
    Thermal & 47.07 & 0.512 & 43.36 & 0.549 &  {44.64} & {0.539} & 43.36 & 0.549\\
    \bottomrule
    \end{tabular}
    }
    \caption{Comparison between different spectra under normal conditions. Models are trained with their corresponding spectrum images from the training set. We test them on the main testing set.  }
    \label{tab:spectra}
\end{table*}

\noindent\textbf{Ablation Studies.} We ablate our proposed mask-guided attention mechanism and the entire spatial transformer to observe their impact. As can be seen in~\cref{tab:results}, the mask-guided attention mechanism notably contributes to TherFormer's performance growth (row (c) vs. (e)), demonstrating its effectiveness in mitigating the effect of background clutter.
Other components in the spatial transformer bring a large performance gain in the RA space (row (e) vs. (d)), confirming the importance of spatial feature enhancement for fine-grained hand joints localization.

\noindent\textbf{Impact of Sequence Length.}
Here we conduct experiments to analyze the impact of the sequence length $T$ on the performance of our baseline method. 
We present the variation in our performance with respect to sequence length in~\cref{fig:length}. All experiments are conducted using the same hyperparameters, except for the $T=16$ model, which uses a smaller batch size due to GPU memory constraints. It can be observed that performance generally improves with increased sequence length, particularly in root-aligned metrics. There is a slight performance decrease in the camera space when the sequence length $T$ is increased to 16 but the performance in the root-aligned space is improved. The general trend underscores the importance of temporal information in enhancing the accuracy of 3D hand pose estimation.

\subsection{Comparison between Spectrum}

\subsubsection{Qualitative Results under Challenging Conditions}
To justify the advantages of using thermal images for egocentric hand pose estimation under challenging scenarios, we conduct a comparison between four spectra on sequences collected in challenging scenarios (\cf\cref{tab:stat}).
To intuitively show the results, we first conduct a qualitative analysis of their performance and present some representative examples in~\cref{fig:challenge}.
We use the same model, \ie, HTT~\cite{wen2023hierarchical}, trained on different spectra
for a fair comparison. 
Please refer to our \href{https://thermohands.github.io/}{\textbf{project page}} for more figure examples and demo videos under challenging conditions.

\noindent\textbf{RGB vs. Thermal.} As can be seen, RGB-based methods fail with gloves wearing (\cf~\cref{fig:challenge} (a-b)) and in darkness (\cf~\cref{fig:challenge} (c-d)). 
Gloves change how human hands look and hide their natural colors and textures. Since RGB algorithms rely on skin's texture and color to identify hand parts and joints, gloves, particularly those with solid colors or textures unlike skin, can interfere with this identification process. Contrarily, as shown in~\cref{fig:challenge} (a-b), thermal imaging methods excel in identifying hands by leveraging the principles of heat conduction through gloves, effectively bypassing the limitations imposed by color and texture variations.

\noindent\textbf{NIR, Depth vs. Thermal.} NIR sensors are significantly disrupted by strong sunlight, affecting both NIR imaging and depth map creation, as shown in \cref{fig:challenge} (e-f). Conversely, thermal imaging is immune to sunlight and outdoor conditions. The temperature difference between hands and their surroundings in the thermal spectrum facilitates effortless identification of the hands, leaving the thermal-based estimator unaffected.
The thermal camera's ability to consistently capture hand features in diverse lighting conditions positions it as a suitable option for future XR applications.

\subsubsection{Quantitative Results under Challenging Conditions}
To provide the numerical results, we evaluate our TherFormer-V models trained for different spectra on our manually annotated sequences (\cf~\cref{setup}) and calculated the quantitative results presented in~\cref{tab:challenge}. As can be seen, thermal imaging-based approaches demonstrate the best performance among different spectra in challenging settings, underscoring thermal imagery’s advantages in difficult lighting conditions and when hands are occluded. Notably, the conclusion drawn from~\cref{tab:challenge} resonate with the visualization in~\cref{fig:challenge}. The RGB spectrum performs poorly in scenarios involving gloves, where the color and texture of human hands are significantly altered. Intense sunlight introduces severe artifacts in NIR images, markedly impairing performance under sun glare.


\subsubsection{Quantitative Results under Normal Conditions} To validate the versatility of thermal-based methods, we also evaluate their performance compared to other spectra (\ie, RGB, depth, NIR and thermal) under normal conditions for general-purpose use cases. We apply different spectral images as  input to multiple baselines methods and report the results in~\cref{tab:spectra}. Not surprisingly, depth image-based methods yield the best performance on average, as they directly utilize the depth information to provide a detailed 3D structure of the hand. The NIR spectrum shows a marginal decrease in performance compared to depth but outperforms both RGB and thermal spectra. This can be attributed to NIR camera's active sensing ability, which leads to more consistent and reliable imaging regardless of the variability of external illuminations (vs. RGB) and temperature (vs. thermal). However, active NIR sensors are prone to interference from external NIR sources like sun glare (\cf~\cref{fig:challenge}). Thermal-based methods, despite using single-channel heat capture, achieves comparable performance to those using RGB images that contain rich color and textures. Such results suggest that thermal images can not only serve as supplements in challenging cases but also can be a viable alternative to other spectra in normal conditions for hand pose estimation.

\section{Limitation and Future Work}
As the first exploration in this field, this work has certain limitations that can guide our future work. First, our dataset primarily focuses on controlled indoor environment and challenging scenarios for RGB/NIR spectrum. Data collection in other challenging scenarios, especially where thermal imaging can be beneficial or limited, is currently scarce. In future works, we plan to expand our dataset beyond the current settings to incorporate a wider range of real-world challenges, including grasping hots objects, reflective surfaces to thermal radiation, ambient heating objects, and more outdoor environments. Paired data should be collected by including and removing
these challenging factors to quantify their impact to thermal imaging-based 3D hand pose estimation.
Second, we only annotate 3D hand pose for our dataset, which limits its usage of evaluation to other tasks relevant to human hands. Further efforts could be annotating the fine-grained hand action splits and hand-object contact~\cite{brahmbhatt2020contactpose}.

\section{Conclusion}

This paper introduces \sysname, the first benchmark for egocentric 3D hand pose estimation using thermal images. \sysname features a multi-spectral, multi-view dataset with automatically annotated 3D hand poses and a novel baseline method, TherFormer, utilizing dual transformer modules for encoding spatio-temporal relationships. We demonstrate near 1cm annotation accuracy, show that TherFormer surpasses existing methods in thermal-based 3D hand pose estimation, and confirm thermal images' effectiveness in challenging lighting and obstruction scenarios. We believe our foundational endeavour could set the stage for further research in thermal-based 3D hand pose estimation and its wide application. 

\section*{Acknowledgments}
This research is partially supported by the Engineering and Physical Sciences Research Council (EPSRC) under the Centre for Doctoral Training in Robotics and Autonomous Systems at the Edinburgh Centre of Robotics (EP/S023208/1) and the grants from the Cisco Research.

\appendix

\section*{Appendix}

The appendix is organized as follows:
\begin{itemize}[label=$\bullet$]
    \setlength{\itemsep}{0pt}
    \setlength{\parsep}{0pt}
    \setlength{\parskip}{0pt}
    \item \cref{dataset} illustrates more details about our hand pose dataset, in the aspects of spectrum coverage, interference avoidance, and dataset distribution.
    \item \cref{annotation} presents more implementation details of our automatic 3D hand pose annotation method.
    \item \cref{network} introduce more architecture design, parameter and loss details of TherFormer. 
\end{itemize}

We also provide more visualization in the format of both images and videos for our data, annotation and results on our \href{https://thermohands.github.io/}{\textbf{project page}}.

\section{Multi-spectral Hand Pose Dataset}\label{dataset}


\noindent\textbf{Spectrum Coverage and Interference Avoidance.} The specification of all sensor frames we collect are shown in~\cref{tab:stat}. Note that we keep the faces of subjects outside of the FoV of the exocentric camera to avoid any personally identifiable information. The multi-spectral data encompasses imaging from the visible spectrum (400-700nm), near-infrared (NIR) spectrum (850 nm $\pm$ 10 nm), and the LWIR spectrum (8-14 $\mu$m). Images captured across different spectra contain unique information and serve varied purposes. For instance, RGB images offer semantic information, facilitating a deeper understanding of human-environment interaction. NIR lights can be actively emitted by our depth cameras to obtain the depth measurements via ToF or structured lighting. 
The LWIR frame, capturing temperature information, readily isolates uniform heat emitters like human hands. On the head-mounted sensor platform, the L515 LiDAR depth camera~\cite{IntelRSL515} emits NIR lasers at a wavelength of 860 nm, which falls outside the thermal camera's receptive range (8-14 $\mu$m), thereby eliminating any potential interference between cameras on the HMSP. Conversely, the exocentric RGB-D camera~\cite{IntelRealSense2023D455} necessitates structured lighting employing NIR at a wavelength identical to the IR emitter on the L515 LiDAR camera~\cite{IntelRSL515}. To prevent interference and image corruption, the exocentric RGB-D camera and the egocentric NIR LiDAR are strategically positioned outside each other's receptive fields during data collection. In the office environment, random heat sources, \eg, servers and chargers, are strategically placed in the background to increase realism and introduce challenging factors into thermal images.

\begin{table*}[!tp]
    \renewcommand\arraystretch{1.0}
    \setlength\tabcolsep{12pt}
    \centering
    \resizebox{2\columnwidth}{!}{%
    \begin{tabular}{@{}lccccccccc@{}}
    \toprule
    & & \multicolumn{3}{c}{Resolution} & \multicolumn{3}{c}{Fov} &\multirow{2}{*}{FPS} \\
        \cmidrule(r){3-5} \cmidrule(r){6-8} 
    \textbf{Sensor frames}  &  Sensor Type & Range & Horizontal & Vertical & Range & Horizontal & Vertical \\
    \midrule
      RGB (ego)  & Intel RS L515~\cite{IntelRSL515} & - & 1280 & 720 & - & 69 & 42 & 30\\
      NIR (ego) & Intel RS L515~\cite{IntelRSL515} & - & 640 & 480 & - & 70 & 55 & 30\\
      Depth (ego) & Intel RS L515~\cite{IntelRSL515} & < 5mm @ 1m & 640 & 480 & 0.25m to 9m & 70 & 55 & 30\\ 
      Thermal (ego) & FLIR Boson 640~\cite{FLIRBoson2024} & $\leq$ 60 mK & 640 & 512 & - & 95 & - & 8.5\\ 
      \midrule
      RGB (exo) & Intel RS D455~\cite{IntelRealSense2023D455} & - & 1280 & 720 & - & 87 & 58 & 30\\ 
      Depth (exo) & Intel RS D455~\cite{IntelRealSense2023D455} & < 2$\%$ at 4m & 848 & 480 & 0.6m to 6m & 87 & 58 & 30\\ 
    \bottomrule
    \end{tabular}
    }
    \caption{The specification of sensor frames captured in data collection.}
    \label{tab:spec}
\end{table*}

\begin{figure*}[tbp!]
    \centering
    \includegraphics[width=0.95\textwidth]{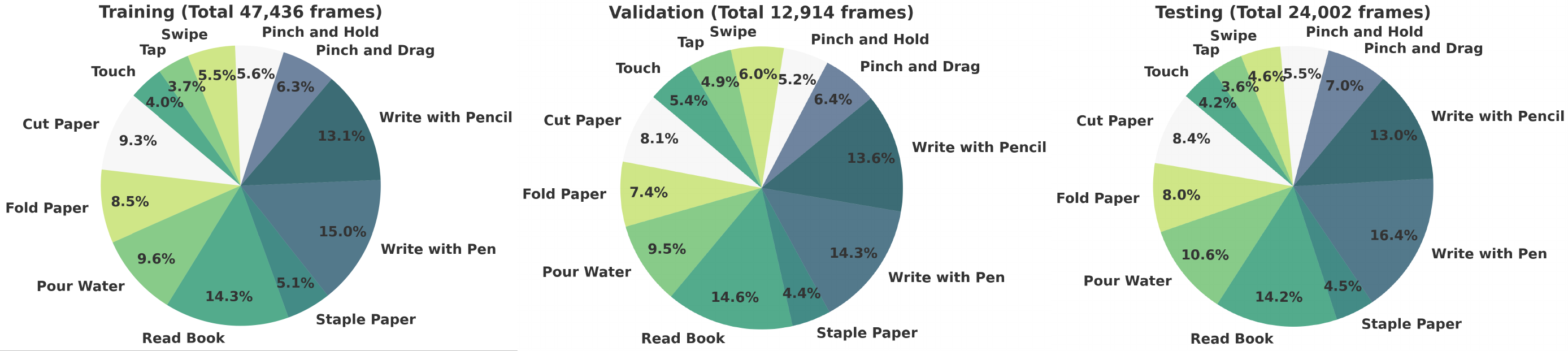}
    \caption{Distribution of data from the main part over hand actions.}
    \label{fig:dist}
\end{figure*}


\noindent\textbf{Dataset Distribution.} We show the data distribution of our main part over hand actions in~\cref{fig:dist}. Among hand-object interaction actions, \emph{write with pen}, \emph{write with pencil} and \emph{read book} have more frames than others due to the complexity of these actions, making them tend to last longer than others. As seven of our participants did not perform the hand-virtual actions, we have fewer frames from them than their hand-object counterparts. Please see our \href{https://thermohands.github.io/}{project page} for more visualization of our collected data.

\section{Dataset Annotation}\label{annotation}

\noindent\textbf{Acquisition of 2D Keypoint and Mask.} We utilize the open-source MediaPipe Hands pipeline~\cite{mediapipe_hands_2020} to infer the 2D hand keypoints from both egocentric and exocentric RGB images. To improve the recall of hand detection, we modify the hyperparameters \texttt{min\_detection\_confidence} from the default 0.5 to 0.1. Sequential RGB images are fed into the MeidaPipe Hands pipeline to obtain the 2D keypoints for two hands. Particularly, we distinguish between two hands according to their relative locations on images. Given 2D keypoints estimated from the task-specific MeidiaPipe Hands model, we employ the versatile Segment-Anything Model (SAM)~\cite{li2020segmenting} to infer the 2D hand masks. To ensure the high quality of the generated 2D hand mask, we utilize the largest version of SAM, \ie,  ViT-L SAM model and prompt it with both the 2D hand keypoints and the bounding boxes defined by them. As a result, we acquire 2D hand keypoints and masks of two hands for each frame. 

\noindent\textbf{3D Keypoint Triangulation.} After inferring the 2D hand keypoints from two views, we can obtain the positions of 3D hand keypoints using triangulation. To implement this, we use the OpenCV function \texttt{cv2.triangulatePoints}~\cite{opencv_library}. 

\noindent\textbf{3D Hand Point Cloud Generation.} To generate 3D hand point cloud, we first index the hand pixels with the 2D hand mask on the depth image and then convert them into 3D points with the camera intrinsic parameters. Points generated from the exocentric view are transformed into the egocentric camera space. By merging points from two viewpoints, we can obtain a dense 3D hand point cloud.

\noindent\textbf{Joint Angle Limitation.} Following~\cite{hampali2020honnotate,kwon2021h2o, liu2022hoi4d}, we optimize the MANO model in the joint angle space instead of the PCA space so that we can constrain the joint angles explicitly. As said in the main paper, the joint angles are limited by a set of empirical boundary values during optimization. Specifically, we use the same parameters of the upper and lower joint angle boundaries as~\cite{kwon2021h2o}. 

\noindent\textbf{Shape Regularization.} To avoid unrealistic hand shape, we also add regularization to the shape parameter $\beta$ when optimizing it for the first frame of each sequence, which can be written as:
\begin{equation}\label{eqn:shape}
    \mathcal{L}_{shape} = \sum_{i=1}^{10} ||\beta_i||
\end{equation}

\noindent\textbf{MANO Fitting.} We utilize the Pytorch version of the MANO layer proposed in~\cite{hasson19_obman}. The Adam optimizer~\cite{kingma2014adam} is used to minimize the optimization objective mentioned in the main paper. To obtain accurate 3D hand pose annotation, we optimize the MANO parameters individually for each frame instead of considering batches of them. For the first frame of each sequence, we run the optimization for 500 iterations with an initial learning rate of as 0.1 decayed by 0.9 for each 50 frames. For the subsequent frames, as we initialize with the results from the last frame, we only optimize for 60 iterations and initialize the learning rate as 0.05 to speed up the convergence. 

\noindent\textbf{Usage Limitations.} Thanks to the well-established RGB-based tools like \cite{kirillov2023segment,mediapipe_hands_2020}, we can extract accurate 2D hand information, which, combined with depth images, enable MANO-based 3D hand pose annotation. However, this approach relies on adequate illumination and minimal occlusion, as hands may appear degraded in RGB or depth images under challenging conditions such as gloves, darkness, and sun glare, and RGB-based tools lack robustness against these factors. Moreover, while our annotation pipeline provides accurate 3D hand pose labels, it is impractical for real-time egocentric estimation due to its reliance on multi-view input, high computational cost, and iterative MANO optimization. In contrast, 3D hand pose estimation methods are designed for real-time inference from a single image view, more suitable for practical deployment.

\section{Baseline Method}\label{network}


\noindent\textbf{Backbone.} The input to our backbone is consecutive thermal images $\mathcal{S} = \{\mathcal{I}_i\}_{i=1}^{T}$ resized to 320$\times$256 after data loading, where $T = 1$ for the single image setting and $T > 1$ for the video setting. We initialize the ResNet-18~\cite{he2016deep} network with the ImageNet-1K~\cite{deng2009imagenet} pre-trained weights for the multi-scale features extraction. The ResNet-18~\cite{he2016deep} backbone returns feature at five levels, with feature sizes as follows given the thermal images as input:
\begin{align*}\small
& \textrm{syntax:}~{Feature ([channel, width, height],\dots)} \\
& {Features ([64, 160, 128], [64, 80, 64], [128, 40, 32],} \\& {[256, 20, 16], [512, 10, 8]) }
\end{align*}

We chose the third level features for future processing considering they carry fine-grained spatial details, which is crucial for accurate mask generation and spatial reasoning in our spatial transformer. 


\noindent\textbf{Mask-guided Spatial Transformer Module.} For spatial feature refinement, we implement a mask-guided attention mechanism that emphasizes regions of interest within the thermal images, specifically focusing on the hands' positions. This approach utilizes predicted masks from a segmentation head with two convolution layers from the mid-level feature to guide the attention mechanism. 

The dimension of the spatial self-attention module we used is 1024, for which we use an additional one-stride convolution to lift the number of channels of the original mid-level feature. Two transformer encoder layers are leveraged and the transformer head number $M$ is set to be 8. The deformable attention mechanism introduced in deformable DETR \cite{zhu2020deformable} is used and the number of sampling points per attention head $K$ is 4:
\begin{equation}\small
\text{DA}(z_q, p_q, x) = \sum_{m=1}^{M} W_m \left[ \sum_{k=1}^{K} A_{mqk} \cdot W'_{m} \left( x(p_q + \Delta p_{mqk}) \right) \right]
\end{equation}
where $x$ is the masked low-level feature, 
$m$ indexes the attention head,
$\Delta p_{mqk}$ and $A_{mqk}$ denote the sampling offset and the learnable attention weight for the $k$-th sampling point in the $m$-th attention head respectively. Particularly,
$A_{mqk}$ is a scalar attention weight lying between 0 and 1, and normalized such that the sum across all $K$ points is 1,
$\Delta p_{mqk}$ is a pair of 2D real numbers representing an unconstrained range.
Noted that 2-D reference point $p_q$ is selected from the hand area derived from the predicted mask and 
$\Delta p_{mqk}$ and $A_{mqk}$ are obtained via linear projection over the query feature $z_q$. In our self-attention, $z_q$ is the added feature for $x$ and the corresponding position encoding. Following deformable DETR \cite{zhu2020deformable},
the first $2M \times K$ channels encode the sampling offsets and the remaining $M \times K$ channels are for the softmax operation to obtain the attention weights.

The output spatial embedding of the deformable mask-guided spatial attention module has the same size as the input feature, which is [1024, 40, 32]. Three convolution layers with a stride of 2 are used to downsample the feature map to [128, 10, 8]. 

\noindent\textbf{Temporal Transformer Module.} This module is crucial for understanding temporal patterns over consecutive frames, utilizing a transformer-based approach. Given per-frame downsample feature map from the spatial transformer module, we flattened them and apply a linear layer to project this spatial embedding to [1,512]. Per-frame spatial feature vectors are then fed to the temporal self-attention, which attends to the feature vector of every frame. The number of attention layers is $2$ and eight heads are used for each attention. The resulting spatio-temporal embedding has the same dimension as the input spatial features, which is [$T$,512].

\noindent\textbf{Pose Regression.} In the pose head, we use the MLP to project the spatio-temporal embedding to the output space and obtain the per-frame 3D joint $\mathcal{J}_i$ for each frame individually. 
\begin{flalign*}
& MLP (512 \rightarrow 512\rightarrow 512 \rightarrow 42\times 3 )
\end{flalign*}



\noindent\textbf{Loss Functions.} Two losses are used for joint learning of hand segmentation and pose estimation during the training process. The L1 hand pose regression loss is defined as: 
\begin{equation}\footnotesize
L_{Hand} = \|P^{2D} - P^{2D}_{gt}\|_1 + \lambda_1\|P^{depth} - P^{depth}_{gt}\|_1 + \lambda_2\|P^{3D} -P^{3D}_{gt}\|_1
\end{equation}
where $P^{3D}$, $P^{3D}_{gt}$,$P^{2D}$, $P^{2D}_{gt}$ are the 3D/2D projection of estimated hand joint positions and ground truth hand joint positions.  $P^{depth}$  and $P^{depth}_{gt}$ are the corresponding depth values for the hand joints. $\lambda_1$ and $\lambda_2$ are the weight parameters set as 100 in our implementation. The hand mask binary cross entropy loss is formulated as follows:

\begin{equation}\small
\begin{aligned}
    L_{Mask} & =  -\frac{1}{WH} \sum_{w=1}^{W} \sum_{h=1}^{H} [w_{pos} \cdot M_{wh} \cdot \log(\hat{M}_{wh}) \\
    &+ w_{neg} \cdot (1 - M_{wh}) \cdot \log(1 - \hat{M}_{wh})]
\end{aligned}
\end{equation}
\normalsize 
where $M_{wh}$ is the binary ground truth label at pixel location $(w,h)$ and $\hat{M}_{wh}$ is the predicted probability for the estimated mask. $w_{pos}$ and  $w_{neg}$ are the weights to cope with the foreground-background imbalance. We set them to $30:1$ during our training.

\noindent\textbf{Training.} We apply the Adam optimizer for our baseline training with an initial learning rate of 1e-4, decayed by a coefficient of 0.95 every epoch. All experiments are conducted using two RTX 3090 GPUs, with a batch size of 16 per GPU. Models are trained for 25 epochs.

\bibliographystyle{unsrt}
\bibliography{egbib}

\end{document}